\pdfoutput=1

\documentclass[11pt]{article}

\usepackage[]{acl}

\usepackage{times}
\usepackage{latexsym}
\usepackage{verbatim}

\usepackage[T1]{fontenc}

\usepackage[utf8]{inputenc}
\usepackage{paralist}
\usepackage{amssymb}
\usepackage{marginnote}

\usepackage{microtype}

\usepackage{inconsolata}
\usepackage{multirow}
\usepackage{booktabs}
\usepackage{graphicx}
\usepackage{framed}
\usepackage{makecell} 
\usepackage{lipsum} 
\usepackage{enumitem}
\usepackage{paralist}
\usepackage{booktabs}

\title{Interpreting Answers to Yes-No Questions\\in Dialogues from Multiple Domains}

 \author{Zijie Wang\textsuperscript{1} \quad
         Farzana Rashid\textsuperscript{2} \quad
         Eduardo Blanco\textsuperscript{1}\\
         \textsuperscript{1}University of Arizona \quad
         \textsuperscript{2}University of North Carolina Asheville\\
         \texttt{\{zijiewang, eduardoblanco\}@arizona.edu} \quad \texttt{frashid@unca.edu}
 }

\begin{document}
\maketitle
\begin{abstract}
People often answer yes-no questions without explicitly saying \emph{yes}, \emph{no}, or similar polar keywords.
Figuring out the meaning of indirect answers is challenging,
even for large language models.
In this paper, we investigate this problem working with dialogues from multiple domains.
We present new benchmarks in three diverse domains: movie scripts, tennis interviews, and airline customer service.
We present an approach grounded on distant supervision and blended training to quickly adapt to a new dialogue domain.
Experimental results show that our approach is never detrimental and yields F1 improvements as high as 11-34\%.
\end{abstract}

\section{Introduction}

While state-of-the-art models obtain results as high as 93\% F1~\cite{zhang2021retrospective}
with question-answering benchmarks such as SQuAD~\cite{rajpurkar-etal-2018-know}, challenges remain.
For example,
natural questions submitted to search engines remain challenging~\cite{kwiatkowski-etal-2019-natural}.
Similarly,
existing models face challenges with
open-ended questions checking for comprehension~\cite{xu-etal-2022-fantastic},
yes-no questions that require deriving an answer (yes or no) from text~\cite{clark-etal-2019-boolq},
false presuppositions and assumptions in the question \cite{yu-etal-2023-crepe,kim-etal-2023-qa},
and negation~\cite{ravichander-etal-2022-condaqa}.


\begin{figure}
\centering
\small
\begin{tabular}{l @{:\ } p{2.6in}}
 \toprule

A$_1$ & I understand, but I noticed that the Fire Marshall is here with you. Is this somehow related to the fire department?	\\
B$_1$ & I really can't give out any information right now at this point.	\\

A$_2$ & Okay. But I do understand that your partner, Leon Jackson's been injured. Is that correct?\\
B$_2$ & He was hurt, but not seriously. He'll be fine.	\\

A$_3$ & Do you have the suspect in custody? \\
B$_3$ & [\ldots{}] is not a good time, okay. Detective Jackson's hurt. He's fine. I've got a Fire Marshall shot, Detective Jackson is hurt but not seriously. \\

\bottomrule
\end{tabular}

\caption{
  Movie dialogue with three yes-no questions (A$_1$, A$_2$, and A$_3$). 
  Answers are indirect as they do not include polar keywords (\emph{yes}, \emph{no}, etc.).
  In B$_1$ and B$_3$, the author declines to answer,
  whereas in B$_2$ the author indirectly answers \emph{no} by
  minimizing the incident
  (\emph{injured} requires loss of function, while \emph{hurt} does not).
}
  \label{t:motivation}
\end{figure}

Many questions in dialogues expect a \emph{yes} or \emph{no} for an answer. 
Yet many follow-up turns answer this kind of questions without explicitly saying \emph{yes}, \emph{no}, or similar polar keywords.
\newcite{hockey1997can} analyze 18 hours of speech~\cite{rossen1997yes} and report that 27\% of questions fall in this category.
Indirect answers to yes-no questions are used to
ask follow-up questions 
or provide explanations for negative answers~\cite{stenstrom1984questions},
prevent incorrect interpretations of direct answers~\cite{hirschberg1985theory},
or show politeness~\cite{brown1978universals}. 

Consider the dialogue from the Movie \emph{15 Minutes} in Figure \ref{t:motivation}.
None of the questions are answered explicitly.
Speaker B declines to answer the first and third questions.
The answer to the first question states that B is not allowed to answer.
The answer to the third question restates information known from previous utterances but provides no answer.
On the other hand, the answer to the second question implicitly denies that Leon was injured by stating that he was (only) not seriously hurt.

This paper tackles the problem of interpreting indirect answers to yes-no questions~(i.e.,
answers that do not contain \emph{yes}, \emph{no}, or other polar keywords).
Our contributions are:\footnote{New benchmarks and code available at \url{https://github.com/wang-zijie/yn-question-multi-domains}}

\begin{compactenum}
  \item Demonstrating that the problem of identifying yes-no questions in dialogues
    can be automated with high precision, even in out-of-domain dialogues;
  \item Creating three new benchmarks~(300 instances each) to evaluate models to interpret answers to yes-no questions in three domains;\\
  \item A methodology using distant supervision to obtain additional (noisy) training data in a new domain with minimal human intervention;
  \item Experimental results showing that blended training with the additional (noisy) data is always beneficial across domains;
  and
  \item Error analysis providing insights into the most difficult indirect answers to interpret correctly.
\end{compactenum}

As our experimental results show, interpreting indirect answers to yes-no questions
is a challenging problem.
In addition, this problem opens the door to several applications.
For example, the work presented here could help dialogue systems avoid conflicts in follow-up turns~\cite{qin-etal-2021-dont}
and alleviate the need for clarification questions~\cite{rao-daume-iii-2018-learning}. 
Further, knowing the interpretations of an indirect answer
could help reveal the intention behind questions~\cite{mirzaei-etal-2023-real}.

\section{Terminology and Existing Corpora}
\label{s:terminology}

We use the term \emph{yes-no question} to refer to a question that expects a \emph{yes} or \emph{no} for an answer.
Answers to yes-no questions may not include \emph{yes}, \emph{no},
or other polar keywords  (e.g., positive: \emph{sure}, \emph{of course}, etc.; negative: \emph{not at all}, \emph{no way}, etc.). 
We refer to answers with and without polar keywords as~\emph{direct} and~\emph{indirect} answers.

We make a distinction between the source of dialogues---who the speakers are and why they communicate. 
We use the term \emph{synthetic} dialogue to refer to dialogues between people who are instructed (and usually paid) to talk about a given topic.
The speakers in synthetic dialogues include crowdworkers. 
We use \emph{genuine dialogue} to refer to naturally-occurring dialogues between people.

Finally, in this paper we work on two problems related to yes-no questions.
Given a dialogue, \textit{identifying} yes-no questions pinpoints where the yes-no questions are.
On the other hand, \textit{interpreting} answers to yes-no questions figures out the underlying meaning of the answer (\emph{yes}, \emph{no}, or \emph{middle}).
Unlike traditional question answering, answers are readily available---the problem is to figure out what the answer means.

\paragraph{Existing Corpora}
We work with several existing dialogue corpora in multiple domains.
In order to identify yes-no questions, we work with the following as training corpora (in-domain):
SWDA~\cite{stolcke-etal-2000-dialogue}, telephone conversations with dialogue act annotations (122k turns);
MRDA~\cite{shriberg-etal-2004-icsi}, meeting transcripts with dialogue act annotations (43k turns);
DailyDialog~\cite{li-etal-2017-dailydialog}, multi-turn conversations written by crowdworkers to simulate human daily conversations (87k turns); 
Friends~\cite{chen-choi-2016-character}, scripts of the TV show (58k turns);
and
MWOZ~\cite{zhu-etal-2020-crosswoz}, task-oriented dialogues written by crowd workers (105k turns).
For evaluation purposes (out-of-domain), we use the following:
Tennis~\cite{liye2016tie}, transcripts of post-match interviews of tennis players (164k turns);
Movie~\cite{danescu-niculescu-mizil-lee-2011-chameleons}, movie transcripts (304k turns);
and Air~\cite{wei-etal-2018-airdialogue}, task-oriented dialogues with topics limited to travel and flights (3,805k turns).

In order to interpret answers to yes-no questions,
we work with the following as training corpora:
Circa~\cite{louis-etal-2020-id}, 34k yes-no questions and indirect answers written by crowdworkers; and
SWDA-IA~\cite{sanagavarapu-etal-2022-disentangling}, 2.5k yes-no questions and indirect answers from the SWDA.
Both corpora include manual annotations of the interpretations of answers.
For evaluation purposes, we use the same corpora than for identifying yes-no questions: Tennis, Movie, and Air.
Specifically, we create new benchmarks (300 questions and indirect answers from each corpus)
and use the rest (questions and direct answers) for training purposes via distant supervision.

\section{Related Works}

Yes-no questions have received considerable attention recently.
BoolQ \cite{clark-etal-2019-boolq} is a collection of 16,000 yes-no questions
and Wikipedia articles
from which answers (\emph{Yes} or \emph{No}) can be derived.
\newcite{sulem-etal-2022-yes} enhance BoolQ with questions that cannot be answered.
Unlike them, in this paper we target yes-no questions in dialogues, which are more open-ended (Figure \ref{t:motivation}) than the fact-seeking questions
(e.g., Has the UK been hit by a hurricane?).
Two recent works target yes-no questions in dialogues~\cite{choi-etal-2018-quac,reddy2019coqa}.
Unlike us, both of them work with synthetic dialogues written by crowdworkers and are constrained to a handful of scenarios.

Yes-no questions in genuine dialogues have been studied before.~\newcite{de-marneffe-etal-2010-good} study 224 yes-no questions including gradable adjectives, 
and~\newcite{de-marneffe-etal-2009-simple} present a typology for 623 yes-no questions from SWDA.
We work with an order of magnitude more data,
several dialogue domains,
and modern learning strategies.

\begin{table*}
  \centering
  \small
  \begin{tabular}{ l rrrrrr} 

\toprule

                        & SWDA & MRDA & DailyDialog & Friends & MWOZ & All  \\ 

\midrule

 genuine dialogue?      & Yes  & Yes &  No  & Yes & No   &  n/a  \\ 
 \# turns               & 122k & 43k &  87k & 58k & 105k &  415k  \\ 

\midrule

 \# yes-no questions    & \\ 
 ~~~~using strict rules             & 1.8k & 1.3k & 8.0k & 2.4k & 16.2k & 29.8k \\
 ~~~~~~~~\# with indirect answers   & 0.0k & 0.0k & 0.0k & 0.0k &  0.0k &  0.0k \\
 ~~~~~~~~precision (in 200 samples) & 1.00 & 1.00 & 1.00 & 1.00 &  1.00 &  1.00 \\
 
 \addlinespace
 
 ~~~~using relaxed rules            & 3.7k & 2.8k & 13.5k & 4.9k & 34.5k & 59.4k \\
 ~~~~~~~~\# with indirect answers   & 1.9k & 1.5k &  5.5k & 2.5k & 18.3k & 29.6k \\
 ~~~~~~~~precision (in 200 samples) & 0.99 & 0.99 &  1.00 & 0.99 &  1.00 & 0.99 \\
 
 \bottomrule
 
\end{tabular}

  \caption{Evaluation of rules to collect yes-no questions. 
    Precision is calculated with a random sample of size 200 for each corpus.
    The relaxed rules yield twice as many yes-no questions~(i.e., twice the relative recall)
    without lowering precision.
    Note that many answers to yes-no questions are \emph{indirect}.} 
  \label{table:rule_results}
  \end{table*}

The work presented here is closest to Circa~\cite{louis-etal-2020-id}, DIRECT~\cite{takayama-etal-2021-direct-direct}, and SWDA-IA~\cite{sanagavarapu-etal-2022-disentangling}.
Unlike us,
Circa works with synthetic yes-no questions and answers without any conversational context.
DIRECT also works with synthetic dialogues. 
SWDA-IA works with telephone conversations from SWDA.
To our knowledge, we are the first to explore yes-no questions in multiple dialogue domains.
We show that existing corpora are beneficial, and more importantly, that combining additional training data obtained via distant supervision in the new dialogue domains brings additional improvements across all domains.

\section{Identifying Yes-No Questions}
\label{s:identifying}

We first tackle the problem of identifying yes-no questions in dialogue.
To our knowledge, previous work on yes-no questions is limited to interpreting the answers.
We first present our rule-based approach to collect yes-no questions. 
Then, we describe how to leverage these rules to build a classifier to automate the task.

\subsection{Collecting Yes-No Questions}
\label{ss:rules}

We define rules to identify yes-no questions in dialogues based on
(a)~dialogue acts if gold annotations are available or (b)~lexical matching.
Our rules look at
(a)~all turns within a dialogue for turns that may contain a yes-no question
and
(b) the next turn to check for direct answers.
We refer to the set of rules that only look at the question as \emph{relaxed rules},
and to the combination of rules that look at the question and answer as \emph{strict rules}.

\paragraph{Corpora with Dialogue Acts Annotations}
For SWDA and MRDA, the only two corpora with dialogue act annotations,
we use these annotations as they indicate yes-no question presence.
Specifically, we refine the list of dialogue acts by \newcite{sanagavarapu-etal-2022-disentangling},
as SWDA and MRDA use different label sets (see Appendix~\ref{a:appendixa}).

\paragraph{Corpora without Dialogue Acts Annotations}
For the other corpora we work with (DailyDialog, Friends, and MWOZ),
we define simple rules that identify yes-no questions with high precision:

The conversation turn:
\begin{compactenum}
\item includes common auxiliary verbs in yes-no questions (%
\textit{do}, \textit{does}, \textit{did},
\textit{don't}, \textit{doesn't}, \textit{didn't}
\textit{is}, \textit{isn't},
\textit{are}, \textit{aren't},
\textit{was}, \textit{wasn't},
\textit{were}, \textit{weren't},
\textit{have}, \textit{haven't},
\textit{has}, \textit{hasn't},
\textit{can}, \textit{can't},
\textit{could}, \textit{couldn't},
\textit{will}, \textit{won't},
\textit{would}, \textit{wouldn't},
\textit{may}, and \textit{might})
and does not include wh-question words (\textit{what}, \textit{when}, \textit{where}, \textit{which}, \textit{who}, \textit{whom}, \textit{whose}, \textit{why}, and \textit{how}); and
\item has more than three tokens and ends in `?'.
\end{compactenum}

\begin{table*}[h!]
  \small
  \centering
  \begin{tabular}{l  rr c rr rr rr rr}
\toprule
& \multicolumn{2}{c}{\multirow{2}{*}{In-Domain}} &&
  \multicolumn{8}{c}{Out-of-Domain} \\ \cmidrule(lr){5-12}
& \multicolumn{2}{c}{} &&
  \multicolumn{2}{c}{Tennis} &
  \multicolumn{2}{c}{Movie} & 
  \multicolumn{2}{c}{Air} &
    \multicolumn{2}{c}{All} \\ \cmidrule(lr){2-3}
                               \cmidrule(lr){5-6}
                               \cmidrule(lr){7-8}
                               \cmidrule(lr){9-10}
                               \cmidrule(lr){11-12}
& \multicolumn{1}{c}{\# k} & \multicolumn{1}{c}{P } &&
  \multicolumn{1}{c}{\# k} & \multicolumn{1}{c}{P} &
  \multicolumn{1}{c}{\# k} & \multicolumn{1}{c}{P} &
  \multicolumn{1}{c}{\# k} & \multicolumn{1}{c}{P} &
  \multicolumn{1}{c}{\# k} & \multicolumn{1}{c}{P} \\
\midrule
Rule-based classifier & \\
~~~strict rules
& 29.8 & 1.00 &&
  23 & 1.00 & 9 & 1.00 

  & 364 & 1.00 & 396 & 1.00 \\
~~~relaxed rules 

& 59.4 & 0.99 &&
  34 & 1.00 & 18 & 0.99 
 
  & 808 & 1.00 & 860 & 1.00 \\
\addlinespace
BERT, distant supervision with&
  \\
~~~strict rules
  & n/a & n/a &&
  40  & 0.99 & 25 & 0.98

  & 826  & 1.00 & 891 & 0.99 \\
~~~relaxed rules 

  & n/a & n/a &&
  42  & 0.99 & 24 & 0.98

  & 825 & 1.00 & 891 & 0.99 \\

\bottomrule
\end{tabular}
  \caption{
    Evaluation of the rule-based and BERT classifiers to identify yes-no questions.
    \emph{In-domain} refers to the corpora used to define the rules (Table \ref{table:rule_results})
      and train the BERT classifier using distant supervision.
    `\# k' stands for \emph{number of yes-no questions identified in thousands}.
    The classifiers with strict and relaxed rules (top block)
    are equally precise, 
    but the latter doubles recall (twice \# k).
    The BERT classifiers are equally precise. }

  \label{t:yn_identification_results}
  \end{table*}
  

Regardless of how questions are identified,
we experiment with an extra rule to check if the next turn is a direct answer.
Here we are not concerned with the interpretation of the answer.
Rather, we consider the subset of yes-no questions that are followed by a direct answer regardless of its interpretation.
We identify direct answers by checking whether the first two sentences in the next turn contain 
\textit{yes}, \textit{yea}, \textit{yup}, \textit{yep}, \textit{yeah}, \textit{sure}, \textit{no}, or \textit{nope}.

Table \ref{table:rule_results} analyzes the outcome of the rules.
We estimate precision using a sample of 200 matches per dialogue domain (total: 1,000),
and use the number of matches (in our case, the number of yes-no questions) to approximate relative recall~\cite{pantel-pennacchiotti-2006-espresso}.
Overall,
The relaxed rules
yield
twice as many yes-no questions than the strict rules (twice relative recall: 29.8k vs. 59.4k matches)
while being equally precise.

\subsection{Classifiers to Identify Yes-No Questions}
\label{ss:identifying}
The rules to identify yes-no questions were iteratively defined using the five corpora in Table~\ref{table:rule_results}: SWDA, MRDA, DailyDialog, Friends and MWOZ.
While the precision is high in these corpora (in-domain),
our goal is to identify yes-no questions in any dialogue (out-of-domain). 
To do so, we evaluate with out-of-domain corpora (Movie, Tennis, and Air) 
with (a) a rule-based classifier
and (b) a classifier trained with the output of the rules using distant supervision.

\paragraph{Rule-based Classifier}
Our first classifier to identify yes-no questions is simple:
run the rules previously defined to identify yes-no questions.
Doing so has the advantage of simplicity.
However, like any other rule-based system, doing so may suffer from low recall as the rules may not generalize.

\paragraph{BERT Classifier}
Our second classifier uses distant supervision.
We use a BERT classifier~\cite{devlin-etal-2019-bert} trained as follows.
We use as positive examples the 59.4k yes-no questions identified with our rules in the training corpora~(Table \ref{table:rule_results}).
As negative examples, we randomly choose 59.4k turns not identified as yes-no questions with the rules.
We use the implementation by Hugging Face \cite{wolf-etal-2020-transformers}.
While any other models could be used, we chose BERT because it demands less computational resources and obtains hard-to-beat results.
Appendix~\ref{a:appendixa} provides additional details.

\begin{table*}[h!]
  \small
  \centering
  \begin{tabular}{ l c rr c rrrr} 
\toprule

  &  & \multicolumn{2}{c}{Existing Benchmarks} & & \multicolumn{3}{c}{Our Benchmarks} \\ 

\cmidrule(lr){3-4} \cmidrule(lr){6-8}

  &  & Circa & SWDA-IA & & Tennis & Movie & Air \\

\midrule

genuine? & & No & Yes & & Yes & Yes & No  \\ 
context? & & No & Yes & & Yes & Yes & Yes \\

\# yes-no questions (train+dev / test) & & 27.4k / 6.8k & 2.0k / 0.5k & & 0 / 300 & 0 / 300 & 0 / 300 \\

~~~~\% answers with interpretation \emph{Yes}    & & 57.1 & 61.9 & & 47.0 & 26.7 & 90.0 \\
~~~~\% answers with interpretation \emph{No}     & & 40.1 & 23.2 & & 18.3 & 18.3 & 3.0  \\
~~~~\% answers with interpretation \emph{Middle} & & 2.8  & 14.9 & & 34.7 & 55.0 & 7.0  \\

\bottomrule
 
\end{tabular}
  \caption{Analysis of corpora to interpret indirect answers to yes-no questions.
    Note that the label distribution (\% of \emph{Yes}, \emph{No}, and \emph{Middle}) is very different in each benchmark. 
    }
  \label{table:classifiet_stat}
  
  \end{table*}

\subsection{Results and Analysis}
\label{ss:identifying_results}

Table \ref{t:yn_identification_results} presents results with the rule-based and BERT-based classifiers.
In-domain refers to the corpora used to define the rules (Table \ref{table:rule_results}, same as \emph{All}).
Out-of-domain includes three additional dialogue corpora
that we will also use to interpret answers to yes-no questions.
We approximate precision using a sample of 200 examples per corpora.

The rule-based classifier obtains almost perfect precision with both in-domain corpora and the three out-of-domain corpora,
regardless of whether we use strict or relaxed rules.
Using relaxed rules, however, obtains twice the amount of yes-no questions in both in-domain and out-of-domain corpora.

Despite it is trained with yes-no questions matching a handful of rules,
the BERT-based classifier identifies many more yes-no questions than the rules themselves.
While the overall benefit looks somewhat low (860k vs. 891k; 3.6\%),
this is mostly due to the small improvement with Air (825k vs. 808k; 2.2\%).
Indeed, in Tennis and Movie the BERT-based classifier identifies 23.5\% and 33.3\% more yes-no questions (42k vs. 34k and 24k vs. 18k).
Note that unlike Tennis and Movie, Air consists exclusively of synthetic dialogues to make travel reservations.
These dialogues are very restrictive; most of the yes-no questions are asked by the speaker acting as the travel agent.
Further, dialogues are scripted and yes-no questions follow very few patterns that can be easily caught with our rules
(e.g., \emph{Do you mean [\ldots]?}, \emph{Would you like a late flight?}).
Surprisingly, there is no difference in training with the output of strict or relaxed rules.

\section{Interpreting Answers to Yes-No Questions}
\label{s:interpreting}

Armed with the highly precise classifier to identify yes-no questions,
we move to interpreting answers to yes-no questions in dialogues from multiple domains.
To our knowledge, there are two publicly available corpora: Circa and SWDA-IA
(Section \ref{s:terminology}).
We aim to explore multiple dialogue domains, so we first create new benchmarks.

\paragraph{Three New Benchmarks}
We create three new benchmarks for evaluation purposes in new domains.
Specifically, we randomly select 300 yes-no questions followed by an indirect answer from each corpus (Tennis, Movie, and Air; 900 total).
Then, we manually annotate the interpretation of the answer using three labels: \emph{Yes}, \emph{No}, or \emph{Middle}.
Our definition of \emph{Yes} includes what Circa and SWDA-IA define as \emph{Probably yes},
which include \emph{sometimes yes} and \emph{yes under certain conditions}.
For example, we annotate \emph{Q: Do you like Mexican food? A: I am fine with tacos if my friends suggest Mexican} with \emph{Yes}.
Similarly, our definition of \emph{No} includes what others define as \emph{Probably no}.
For example, we annotate \emph{Q: Do you want to go out for dinner? A: I have a deadline and I may skip dinner} with \emph{No}.
On the other hand, our definition of \emph{Middle} follows previous work:
we choose it when the answer does not lean toward \emph{yes} or \emph{no}.

Table \ref{table:classifiet_stat} summarizes all the benchmarks available.
Training data is only available for the two existing benchmarks.
Note that the label frequency is very different between existing corpora and our new benchmarks.
We argue that not artificially balancing the benchmarks is sound.
Indeed, domain adaptation is not only about working with language from other domains,
but also accounting for label distribution shifts.
Tennis and Movie have many more answers to yes-no questions whose interpretation is \emph{Middle} compared to Air 
(34.7\% and 55.0\% vs. 7.0\%) and existing benchmarks (2.8\% and 14.9\%).
We also observe that our benchmarks have fewer answers whose interpretation is \emph{No}.
Most answers in Air are interpreted as \emph{Yes}; as discussed before most questions in Air come from a travel agent confirming travel arrangements rather than open-ended conversations.
The substantial differences in the distribution of interpretations across existing and our benchmarks
indicate that transfer learning across these domains might be challenging.
As we shall see, however, we benefit from using distant supervision with all the dialogue corpora.

\paragraph{Inter-Annotator Agreements}
The three benchmarks were annotated in-house by two graduate students.
Inter-annotator agreements (Cohen's $\kappa$) for Tennis, Movie, and Air are 0.68, 0.69, and 0.66 respectively.
These coefficients indicate \emph{substantial} agreement \cite{artstein-poesio-2008-survey};
above 0.8 would be (nearly) perfect.
87.0\% of disagreements are between
(a)~\emph{Yes} or \emph{No}
and
(b)~\emph{Middle},
while only 13.0\% are between \emph{Yes} and \emph{No}.
These percentages suggest that most disagreements are minor.
After annotating individually, the annotators discussed the disagreements in order to adjudicate them and create the final ground truth.
We refer the reader to Appendix~\ref{a:appendixb} for more details.

\subsection{Model Training Strategies}
\label{ss:strategies}
We follow three strategies to build models to interpret answers to yes-no questions.
The differences are which corpora we train with and the training procedure to combine the training corpora.
All strategies start with the off-the-shelf RoBERTa transformer~\cite{liu2019roberta}
released by  Hugging Face~\cite{wolf-etal-2020-transformers}.
This problem is harder than identifying yes-no questions,
and we found it beneficial to use RoBERTa instead of BERT.
We also experiment with BART~\cite{lewis-etal-2020-bart},
however, RoBERTa 
outperforms BART on most benchmarks---the only exception is Air.
Appendix~\ref{a:appendixc} details the results with BART.
All hyperparameters were tuned with the train and validation splits;
we refer the reader to Appendix~\ref{a:appendixc} for details.

The first strategy is to fine-tune a RoBERTa classifier with the existing benchmarks (Circa and SWDA-IA)---the only ground truth available for training purposes to interpret yes-no questions.
The other two strategies also fine-tune a RoBERTa classifier, but combine training data from
(a)~existing benchmarks
and
(b)~additional instances from the same corpora we created our benchmarks with.
These additional instances were obtained using distant supervision as detailed below.
Crucially, obtaining them does not require human involvement after generic patterns applicable to any dialogue corpora are defined.
The second and third strategies differ in the fine-tuning methodology.
The former merges the training data and proceeds to fine-tune with the combination.
The latter uses blended training to phase out the training data from existing corpora as detailed below.
\begin{table*}[h!]
  \small
  \centering
  
\begin{tabular}{ l rr c rr@{}l r@{}l } 
 
 \toprule
 
   & \multicolumn{2}{c}{Existing Benchmarks} & & \multicolumn{5}{c}{Our Benckmarks} \\ 

\cmidrule(lr){2-3} \cmidrule(lr){5-9}
   
   & Circa & SWDA-IA & & Tennis & Movie && Air   \\ 
 
\midrule

Majority Baseline & 0.43 & 0.32 & & 0.34 & 0.36 && 0.84 \\

\midrule

\multicolumn{7}{l}{RoBERTa, training with} \\

~~~~Circa         & 0.93 & 0.51 & & 0.40 & 0.34 && 0.84 \\
~~~~SwDA-IA       & 0.69 & 0.63 & & 0.42 & 0.37 && 0.63 \\
~~~~Circa + SwDA-IA & 0.93 & 0.68 & & 0.52 & 0.37 && 0.84 \\
\midrule

\multicolumn{7}{l}{RoBERTa, training with} \\

\multicolumn{7}{l}{~~~in-domain instances and}\\

~~~~~~Circa         & n/a & n/a  & & 0.43 & 0.24 && 0.85 \\
~~~~~~SwDA-IA       & n/a & n/a  & & 0.41 & 0.31 && 0.84 \\
~~~~~~Circa + SwDA-IA & n/a & n/a  & & 0.48 & 0.34 && 0.85 \\

~~~all additional instances and & \\
~~~~~~Circa         & 0.93  & 0.56  & & 0.44  & 0.24 && 0.82  \\
~~~~~~SwDA-IA       & 0.74  & 0.67  & & 0.51  & 0.28 && 0.82  \\
~~~~~~Circa + SwDA-IA & 0.93  & 0.70  & & 0.53  & 0.42 && 0.84  \\
\midrule

RoBERTa, blended training with\\

\multicolumn{7}{l}{~~~in-domain instances and}\\

~~~~~~Circa         & n/a & n/a & & 0.37 & 0.31 && 0.83 \\
~~~~~~SwDA-IA       & n/a & n/a & & 0.48 & 0.36 && 0.84 \\
~~~~~~Circa + SwDA-IA & n/a & n/a & & 0.57 & 0.39 && 0.84 \\

~~~all additional instances and&\\

~~~~~~Circa         & 0.93 & 0.54 & & 0.42 & 0.44 && 0.82 \\
~~~~~~SwDA-IA       & 0.71 & 0.64 & & 0.50 & 0.40 && 0.85 \\
~~~~~~Circa + SwDA-IA & 0.93 & 0.68 & & 0.59 & 0.49&$^\ast$ & 0.86&$^\ast$ \\

\bottomrule
\end{tabular}
  \caption{Results (F1) interpreting answers to yes-no questions.
   \emph{In-domain} refers to additional training instances from the \textit{same} domain we evaluate with.
   \emph{Air} is heavily unbalanced (Table \ref{table:classifiet_stat}) and limited to airline bookings;
     no model substantially outperforms the majority baseline.
   Training with the proposed distant supervision approach is
   (a)~never detrimental if training data in the same domain is available (Existing Benchmarks)
   and
   (b)~always beneficial otherwise (Our Benchmarks).
   The improvements on Movie and Air are statistically significant (McNemar's test~\cite{mcnemar_note_1947}, p < 0.05; indicated with an asterisk).
  }
  \label{t:interpretation_results}
  \end{table*}
  
\paragraph{Blended Training}
We adopt the method by \citet{shnarch-etal-2018-will} to blend training data from existing corpora (Circa and SWDA-IA)
and the additional annotations obtained with distant supervision.
The blending process consists of two phases:
\textit{m} blending epochs using all the additional annotations and a fraction of the training instances from existing corpora,
and then \textit{n} epochs only using all the additional annotations.
The intuition is that existing corpora provide a good base to interpret answers to yes-no questions,
but that it is beneficial to use instances closer to the domain we evaluate with as training progresses.
In each blending epoch, the fraction of instances from existing corpora are fed randomly to the network. 
The blending factor $\alpha \in [0, 1]$ determines the fraction of instances from existing corpora to consider.
The first blending epoch trains with all of them,
and 
the ratio to phase out in each epoch is determined by $\alpha$.
The blending hyperparameters ($\alpha$, $m$, and $n$)
are tuned like any other hyperparameter~(see Appendix~\ref{a:appendixc}).

\paragraph{Distant Supervision}
The goal of distant supervision is to explore whether considering additional instances automatically labeled in the new dialogue domains is beneficial.
Given the high precision of the patterns to identify yes-no questions (Section~\ref{s:identifying}),
using the strict rules and matching \emph{yes} and \emph{no} keywords to their corresponding answers is worth exploring.
The aim of these patterns ought to be as precise as possible.
Disregarding many yes-no questions and answers (i.e., low recall)
is acceptable as the large amount of unannotated dialogue corpora still allows us to automatically label many instances and use them to train models.
We consider the same patterns from Section~\ref{s:identifying}.
The keywords to label an answer as
\emph{Yes} (\textit{yes}, \textit{yea}, \textit{yup}, \textit{yep}, \textit{yeah}, \textit{sure})
or
\emph{No} (\textit{no}, or \textit{nope})
are limited.
However, we found that adding other keywords leads to unnecessary noise.
For example, \emph{sure} appears at first sight to be a good keyword for \emph{Yes},
although it often is not (e.g., \emph{Q: Do you like Mexican food?} \emph{A: Sure, if I run out of everything else I will eat it.}).

Distant supervision in the three new dialogue domains yields
380k instances for training purposes (Tennis: 19,055, Movie: 6,250, and Air: 355,549).
We balanced the datasets before model training.

\subsection{Experimental Results} 
\label{ssec:experiment_results}

We present the results in Table \ref{t:interpretation_results}.
Air is unbalanced (Table \ref{table:classifiet_stat}, Yes: 90.0\%), and all models obtain similar results than the majority baseline.

Let us first discuss the results training only with existing corpora (second block; Circa, SWDA-IA, or both).
Synthetic yes-no questions and answers are much easier to interpret (Circa: 0.93, Air: 0.84, both F1-score) than those coming from genuine dialogues (F1: 0.37--0.68),
although out-of-domain evaluation shows that training with existing corpora outperforms the majority baselines with Tennis~(F1: 0.34 vs. 0.52) and Movie (F1: 0.36 vs. 0.37).

Second, training strategies combining existing training data and the additional instances obtained via distant supervision are beneficial.
In particular, it is beneficial to consider \emph{all} instances (Tennis, Movie, and Air) regardless of which domain we evaluate with.
Finally, we observe that blending (fourth block; 0.49--0.86) yields better results than merging the additional annotations (third block; 0.42--0.85).
Most importantly, for two benchmarks~(Movie and Air), the improvements are 
statistically significant (McNemar's test~\cite{mcnemar_note_1947}, p~< 0.05) when compared to training with the existing data.
Thus the proposed distant supervision is successful at adapting a model to interpret yes-no questions to new domains. 
This is true across all labels despite distant supervision only identifies additional training instances with 
\emph{Yes} and \emph{No} interpretations. Appendix~\ref{a:appendixc} provides additional results (F1 score) per label and more metrics (Precision, Recall, and F1 score)
that complement Table~\ref{t:interpretation_results}.

\begin{table*}
  \centering
  \small
  

\begin{tabular}{l r c@{}c@{}c l r@{, }l }
    \toprule
    Error Type & \% & T & M & S & Example & G & P\\ \midrule
    
    Unresponsive answer & 18 & \checkmark & \checkmark & \checkmark & 
    Q: Do you think it is a little early? &
    Middle & Yes \\
    &&&&& A: I brought you something \ldots From the library. \\

    Answer has question & 13 & \checkmark & \checkmark & \checkmark &
    Q: Really?  Do you have the money with you? &
    Middle & Yes \\
    &&&&& A: Do you have the material? \\
    
    Intricate Answer & 7 & \checkmark & \checkmark & \checkmark &
    Q: [\ldots] do you think the English players have it easier? & 
    Middle & Yes \\
    &&&&& A: [335 tokens] There was a lot of talk about the lack [\ldots] \\ 
    
    Question has negation & 5 & \checkmark & \checkmark & \checkmark &
    Q: Don't you like cats? &
    Yes & No \\
    &&&&& A:  Well, I like cats. This, this cat is a, uh, more like a dog. \\ \midrule

    Polar distractor in answer & 18 & \checkmark && \checkmark &
    Q: You may be familiar with [\ldots]. Have you ever, &
    No & Yes \\
    &&&&& A: [\ldots], you can tell me a little bit more about it [\ldots] \\ 
    
    
    Short question or answer & 13 && \checkmark & \checkmark  &
    Q: Were you &
    Middle & Yes \\
    &&&&& A: Really is. \\
    
    Confrontational answer & 6 & \checkmark & \checkmark &  &
    Q: Did my father tell you not to talk about it? &
    Middle & Yes \\
    &&&&& A: Come on. you brought it up. \\
    
    Uninformative answer & 5 & \checkmark && \checkmark  & 
    Q: Is it twenty percent? &
    Middle & No \\
    &&&&& A: I, I have no idea, I just. My dad does it all for me. \\
    
    External knowledge & 5 && \checkmark & \checkmark  & 
    Q: You're from L.A., huh? &
    No & Middle \\
    &&&&& A: New York. \\ 
    
    Condition or contrast & 4 & \checkmark && \checkmark  &
    Q: Is he able to, uh, still do everything himself pretty well? &
    No & Yes \\
    &&&&& A: Well, he was until this operation. He has arthritis. \\ \bottomrule
    
    \end{tabular}
  \caption{
    Most common error types in \emph{T}ennis, \emph{M}ovie, and \emph{S}WDA-IA with our best model.
    Percentages are the average in the corpora where the error was observed, indicated with checkmarks.
    The last column indicates the \emph{G}old and (wrong) \emph{P}redictions.
  } 
  \label{t:errors}
  \end{table*}

\subsection{Error Analysis}
\label{ss:error_analysis}
We also conduct an error analysis to identify the most common error types made by the best-performing model
(i.e., bottom row in Table \ref{t:interpretation_results}).
We analyze 100 errors from the three dialogue domains our best model makes the most errors with
SWDA-IA~(F1:~0.68), Tennis~(F1:~0.59), and Movie~(F1:~0.49).
Note that we make few errors with Circa and Air (F1:~0.93 and 0.86).

We identify four frequent error types
across the three dialogue domains (first block in Table \ref{t:errors}).
First, unresponsive answers should almost always be interpreted as \emph{Middle} as they do not address the question,
yet the model routinely (18\% of errors) mispredicts \emph{Yes} or \emph{No}.
Similar mispredictions occur for a specific kind of unresponsive answer: answering with a question~(13\%).
Intricate, long answers account for 7\% of errors.
In the example, the answer has 335 tokens; it gives background and explanations but it never addresses the questions (Gold: \emph{middle}).
Note that in Tennis interviews, most of the conversation turns are rather long.
Lastly, we found that 5\% of errors in all three dialogue domains occur when the question has a negation---regardless of the answer.

We also identify six error types in at least two of the dialogue domains.
In Tennis and SWDA-IA, polar distractors (i.e., \emph{yes} or \emph{no} indicators in an answer whose interpretation is \emph{No} or \emph{Yes})
account for 18\% of errors.
In the examples, the model is misguided by \emph{can}, which indicates \emph{Yes} despite the answer ought to be interpreted as \emph{No}.
Extremely short answers are somewhat common in Movie and SWDA-IA and account for 13\% of errors.
We define confrontational and uninformative answers as answers that are hostile towards the author of the questions and avoid providing an answer while not changing the topic of conversation (unlike unresponsive answers, which are discussed above).
Uninformative answers are always to be interpreted as \emph{middle}.
Finally, we also identify that interpreting answers sometimes requires external knowledge such as world and commonsense knowledge (5\% of errors in Movie and SWDA-IA).
In the example, being from \emph{New York} means someone is not from \emph{L.A.}; however, being from \emph{Hollywood} would mean the opposite.
Finally, we identify conditions and contrasts---even in short answers---are present in 4\% of errors in Tennis and SWDA-IA.
In the example, the answer states that \emph{he was able to do everything himself until the operation},
implying that he is not able anymore. Thus, the ground truth is \emph{No}.
The model is unable to see the contrast between the past and current situation.

\begin{table}[h!]
  \small
  \centering
  \begin{tabular}{lrrr} 

    \toprule
    
      & Tennis & Movie & Air\\ 
      \midrule
    RoBERTa, training with\\
    ~~~~additional instances and  \\
    ~~~~~~Circa             & 0.41 & 0.24 & 0.84 \\
    ~~~~~~SWDA-IA           & 0.34 & 0.27 & 0.83 \\
    ~~~~~~Circa + SWDA-IA   & 0.55 & 0.42 & 0.85 \\
    
    \midrule
    RoBERTa, blended with\\
    ~~~~additional instances and  \\
    ~~~~~~Circa             & 0.48 & 0.42 & 0.84 \\
    ~~~~~~SWDA-IA           & 0.41 & 0.29 & 0.85 \\
    ~~~~~~Circa + SWDA-IA   & 0.60 & 0.48 & 0.86 \\
    
    \bottomrule
       
\end{tabular}
  \caption{Results (F1) obtained with RoBERTa trained with a set of additional instances 
  that is the same size as the in-domain instances. The results remain similar compared to the model trained with
  \textit{all additional instances} (Table~\ref{t:interpretation_results}), demonstrating that the performance gains
  are mostly due to training instances from the new domains rather than just more training instances.}
  \label{t:ablation}
\end{table}

\subsection{Ablation Study: More Instances or Cross-Domain Instances?}
To further understand the source of the performance gains---whether they are due to more training 
instances or having cross-domain instances---we conduct an ablation study. This involves training a RoBERTa model with 
additional instances that are equivalent in size to the in-domain instances for each benchmark. 
Table~\ref{t:ablation} presents the results. 

For the second training strategy (Section~\ref{ss:strategies}), 
the results are never detrimental compared to 
training with all additional instances (Table~\ref{t:interpretation_results}, Tennis: 0.53 vs. 0.55, Movie: unchanged 0.42, and Air: 0.84 vs. 0.85), 
demonstrating that the performance gains are mostly from the cross-domain instances. 
A similar trend is observed for the third training strategy (blending; Tennis: 0.59 vs. 0.60, Movie: 0.49 vs. 0.48, and Air: unchanged 0.86).

\subsection{A Note on Large language Models} 
\label{ssec:llm_results}
Recent works have shown that prompting with large language models achieves better results in many tasks~\cite{mishra-etal-2022-reframing} compared to supervised approaches using substantially smaller models.
This is not the case with the problem of interpreting indirect answers to yes-no questions.

We explore whether large language models outperform our RoBERTa-based classifier at interpreting indirect answers to yes-no questions.
Specifically, we experiment with three LLM models: GPT-3.5~\cite{brown2020language}, Alpaca-7B~\cite{alpaca}, and Llama 2-7B~\cite{touvron2023llama}.
We manually map the models' output to our interpretations of indirect answers (\textit{Yes}, \textit{No}, or \textit{Middle}) for evaluation purposes.

We evaluate with GPT-3.5 using Microsoft Azure API. 
For Alpaca and Llama, we host them locally. 
However, we are only able to evaluate them in up to 4-shot prompting because of resource limitations.
Table~\ref{t:llm_results} shows the results. 
In general, 4-shot prompting yields improvements compared to 0-shot prompting.
Surprisingly, GPT-3.5 obtains worse results than the other two models with Movie and Air despite it being a much larger model.
We hypothesize the higher results with Tennis are due to GPT-3.5's better performance with longer texts.
Tennis has on average much longer answers than the other two benchmarks.
Most importantly, all three models fail to match our best results.

To further investigate the reason behind the poor performance, we conduct an error analysis with the results 
obtained with GPT-3.5 using 4-shot prompts. 
We calculate the error distributions by gold label and (wrong) predictions. In addition, we list a few examples.
The results can be found in Appendix~\ref{a:appendixc}, which also contains details about the prompts and experimental setup.

\begin{table}
  \small
  \centering
  \begin{tabular}{lrrr} 

    \toprule
    
      & Tennis & Movie & Air\\ 
      \midrule
    \multicolumn{4}{c}{0-shot}\\
    \midrule
    GPT-3.5       & 0.39 & 0.29  & 0.23 \\
    Alpaca-7B     & 0.35 & 0.22  & 0.28 \\
    Llama 2-7B    & 0.32 & 0.19  & 0.32 \\
    \midrule
    \multicolumn{4}{c}{4-shot}\\
    \midrule
    GPT-3.5       & 0.50 & 0.31  & 0.59 \\
    Alpaca-7B     & 0.43 & 0.40  & 0.77 \\
    Llama 2-7B    & 0.33 & 0.21  & 0.73 \\  
    \midrule
    Our best model    & 0.59 & 0.49  & 0.86\\   
    \bottomrule
       
\end{tabular}
  \caption{Results (F1) obtained with GPT-3.5, Alpaca-7B, and Llama 2-7B. We evaluate them 
  with the test split (240 instances per benchmark) in 0-shot and 4-shot prompts. 
  Despite their much larger model size, none of them outperforms our best model.}
  \label{t:llm_results}
  \end{table}

\section{Conclusions}
Indirect answers to yes-no questions in dialogue are common.
In this paper, we have presented an approach to identify yes-no questions in dialogues~(distant supervision and BERT),
and more importantly,
to interpret indirect answers to yes-no questions.
Experimental results show for the first time that the identification problem is rather simple.
The second problem, on the other hand, remains challenging---F1 scores are 0.49 and 0.59 with Movies and Tennis.
These results lead to the conclusion that synthetic dialogues may not be representative of more open-ended conversations.

Crucially, we have shown that distant supervision to obtain additional examples with \emph{direct} \emph{Yes} and \emph{No} answers is beneficial to interpret \emph{indirect} answers.
Indeed, combining the additional instances with blended training
is never detrimental and yields substantial improvements with our new, out-of-domain benchmarks (Tennis, Movie, and Air).
In other words, the proposed methodology can be used to adapt to new domains without requiring substantial human involvement,
unlike annotating additional examples. 

Our future plans include addressing the most common errors.
In particular, we believe that exploring dialogue coherence and pretrained language models customized to the dialogue domain are two lines of research worth exploring.


\section*{Limitations}

We adopt distant supervision to obtain additional data for training purposes. However, we only design rules to identify yes-no questions with direct answers, 
which means the extra (and noisy) training instances only have interpretations \emph{yes} and \emph{no}---%
there are no additional instances with \emph{middle} interpretations.
We notice that in some cases,
the improvements in results are mostly brought by better results predicting indirect answers that are labeled \emph{yes} and \emph{no}, 
and only to a smaller degree by those labeled \emph{middle}.
Detecting hesitation or non-answers (or simply answers who indicate 50/50) could be critical in some domains and our distant supervision approach
does not provide much benefit with \emph{middle}.

We annotate our new benchmarks with 3 labels~(\emph{yes}, \emph{no} and \emph{middle}).
Some previous works use finer-grained label sets to interpret answers to yes-no questions.
For example, \newcite{sanagavarapu-etal-2022-disentangling} use five labels including \emph{Probably yes} and \emph{Probably no},
and \newcite{louis-etal-2020-id} use nine labels including \emph{Sometime yes}, \emph{In the middle}, or \emph{I am not sure how to interpret [the answer to the question]} 
(only 3.6\% of the answers receive these three labels, however).
Considering that
(a)~there is no universal agreement about the possible ways to interpret answers to yes-no questions,
(b)~other works also use three labels~\cite{sulem-etal-2022-yes},
and
(c)~it is unclear which interpretations may be more useful in a real-world application,
we argue that three labels are as sound as five or nine---or at least not worse.

We tune several hyperparameters
(including the blending factor $\alpha$ and the amount of additional instances to train; third and fourth block in Table \ref{t:interpretation_results})
with the train and development splits,
and report results with the test set.
The results are taken from the output of one run.
We acknowledge that the average of multiple runs (e.g., 10) would be more reliable,
but they also require much more computational resources (literally, 10 more times).


\section*{Ethics Statement}


\paragraph{Data biases.}
Our work only focuses on yes-no questions and answers in English dialogues.
Researchers~\cite{lafford2004effect} have shown that other languages (e.g., Spanish) tend to communicate more directly,
thus the vagueness of indirect answers may not be as big of a problem.
Our future plans include targeting the same problem in multiple languages.

\paragraph{Data sources and collection.}
We collect the Movie, Tennis, Friends and SWDA datasets from Convokit~\cite{chang-etal-2020-convokit};
and samples from MWOZ, under MIT license.
Samples from MRDA are used under GPL-3.0 license.
Samples from DailyDialog are used under CC BY-NC-SA 4.0.
Samples from AirDialogue are used under Apache License 2.0.

\section*{Acknowledgements}
We thank the reviewers for their insightful comments.~We also thank the Chameleon platform~\cite{keahey2020lessons}
for providing computational resources. The Accelerating Foundation Models Research program by Microsoft provided Azure credits to conduct this research.

This material is based upon work supported
by the National Science Foundation under Grant
No. 2310334. Any opinions, findings, and conclusions or recommendations expressed in this material are those of the authors and do not necessarily
reflect the views of the NSF.
\bibliography{anthology}


\appendix
\section{Additional Details to Identify Yes-No Questions}
\label{a:appendixa}

\paragraph{Dialogue Act Labels to Select Yes-No Questions}
The process to select yes-no questions uses dialogue act labels if available (Section~\ref{s:identifying}).

The list of dialogue acts and their descriptions for SWDA is as follows:

\begin{compactitem}
  \item \texttt{qh}: Rhetorical question
  \item \texttt{qy}: Yes-no question
  \item \texttt{qy$\land{}$d}: Declarative yes-no question
  \item \texttt{$\land{}$g}: Tag-Question
  \item \texttt{qy$\land{}$t}: Yes-no question about task
  \item \texttt{qy$\land{}$r}: Yes-no question repeat self
  \item \texttt{qy$\land{}$m}: Yes-no question mimic other
  \item \texttt{qy$\land{}$h}: Question in response to a question
  \item \texttt{qy$\land{}$c}: Yes-no question about communication
  \item \texttt{qy$\land{}$2}: Yes-no question collaborative completion
  \item \texttt{qy($\land{}$q)}: Yes-no question quoted material
  \item \texttt{qy$\land{}$g}: Yes-no question tag-question
  \item \texttt{qy$\land{}$g$\land{}$t}: Yes-no question tag-question about task
  \item \texttt{qy$\land{}$g$\land{}$r}: Yes-no question tag-question repeat self
  \item \texttt{qy$\land{}$g$\land{}$c}: Yes-no question tag-question about communication
  \item \texttt{qy$\land{}$d$\land{}$t}: Declarative yes-no question about task
  \item \texttt{qy$\land{}$d$\land{}$r}: Declarative yes-no question repeat self
  \item \texttt{qy$\land{}$d$\land{}$m}: Declarative yes-no question mimic other
  \item \texttt{qy$\land{}$d$\land{}$h}: Declarative yes-no question in response to a question
  \item \texttt{qy$\land{}$d$\land{}$c}: Declarative yes-no question about communication
  \item \texttt{qy$\land{}$d($\land{}$q)}: Declarative yes-no question quoted material
  \item \texttt{qy$\land{}$c$\land{}$r}: Yes-no question about-communication repeat self
  \end{compactitem}

The list of dialogue acts and their descriptions for MRDA is as follows (this corpus includes fewer dialogue act labels):

\begin{compactitem}
  \item \texttt{qy}: Yes-no question
  \item \texttt{g}: Tag-question	
\end{compactitem}

\paragraph{Details and Hyperparameters for BERT-based Classifier}
Referring to Section~\ref{ss:identifying}, we use an off-the-shelf BERT-base model 
(110M parameters) from Hugging Face~\cite{wolf-etal-2020-transformers} to train a classifier
that identifies yes-no questions.
We run the experiments on a single NVIDIA Tesla V100 (32GB) GPU. 
It takes approximately 5 minutes to train 1 epoch.
Table~\ref{table:hyper_indentify} shows the hyperparameters that yield the highest accuracy in identifying yes-no questions.

\begin{table}
  \small
  \centering
  \begin{tabular}{ l r r} 

    \toprule
    
    Hyperparameters   \\ 
    
    \midrule
    Maximum Epochs    & 5     \\      
    Batch Size        & 32      \\
    Optimizer         & AdamW  \\ 
    Learning rate     & 5e-5   \\

        \bottomrule
       
       \end{tabular}
  \caption{Tuned hyperparameters for experiments to identify yes-no questions with BERT-base model.}
  \label{table:hyper_indentify}
  
  \end{table}

\section{Additional Details about Benchmark Annotations}
\label{a:appendixb}
\paragraph{Annotation Instructions}

We conduct manual annotations to obtain ground truth interpretations (i.e., gold labels) for indirect answers to yes-no questions. 
We adopt three labels: \emph{Yes}, \emph{No}, and \emph{Middle (Unknown)}, and we define them as follows for annotations: 
\begin{itemize}
\item \emph{Yes}: The answer shows (or implies) \emph{yes}, \emph{yes} under certain conditions or constraints (probably yes). 
\item \emph{No}: The answer shows (or implies) \emph{no}, \emph{no} under certain conditions or constraints (probably no), conveys negative sentiment, or provides arguments for \emph{no}.
\item \emph{Middle (Unknown)}: The answer is unresponsive (e.g., changes the topic) or uninformative (e.g., ``I don't know'' answer). It should imply or lean towards neither Yes nor No.
\end{itemize}

\paragraph{Annotator Demographics}

Two annotators including a female and a male are recruited for the dataset annotation. 
Their ages range from 26 to 30 years old. Both of them are from Asia and with a graduate degree in Computer Science.

\paragraph{Annotation Agreements}
\begin{figure}
  \centering
  \small
  \includegraphics[width=0.45\textwidth]{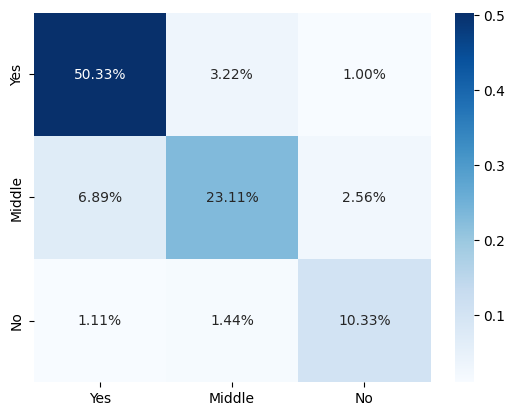}
  \caption{
   Heatmap of the inter-annotator agreements. The percentages are the average of three benchmarks (total: 900 instances). 
   Most disagreements are between (1) \textit{Yes} or \textit{No} and (2) \textit{Middle}.
   }
    \label{f:heatmap}
\end{figure}

Figure~\ref{f:heatmap} shows the percentages of disagreements between the annotators.
Most disagreements are minor (between (a) Yes or No and (b) Middle).
Recall that Cohen's $\kappa$ inter-annotator agreements are between 0.66 to 0.69.

\section{Additional Details to Interpret Indirect Answers}
\label{a:appendixc}
\paragraph{Details and Hyperparameters}
Referring to Section~\ref{ssec:experiment_results}, we use an off-the-shelf RoBERTa-base model (125M parameters) from Hugging Face~\cite{wolf-etal-2020-transformers}. We run the experiments on a single
NVIDIA Tesla V100 (32GB) GPU. Depending on the sizes of the training datasets, it takes approximately 10 minutes to an hour to train one epoch.
Table~\ref{table:hyper_interpret} shows the hyperparameters that lead to the highest F1 score in interpreting indirect answers.

\begin{table*}[h]
  \small
  \centering
  \begin{tabular}{ l r r} 

\toprule

Hyperparameters  & Training with extra data & Blended training\\ 

\midrule
Maximum Epochs    & 30    & 20 \\
Warmup steps      & 500   & 200 \\  
Batch Size        & 32    & 16  \\
Optimizer         & AdamW & AdamW \\ 
Learning rate     & 2e-5  & 2e-5 \\
Weight decay      & 1e-2  & 1e-2 \\
Gradient clipping & 1.0   & 1.0  \\

    \bottomrule
   
   \end{tabular}
  \caption{Tuned hyperparameters for experiments to interpret indirect answers with RoBERTa-base model.}
  \label{table:hyper_interpret}
  
  \end{table*}

  We also tune the number of additional training instances and blending factor $\alpha$ as other parameters. 
  We choose the number of additional training instances from (2k, 5k, 10k)
  (or all instances if they are less than the number), and
  the $\alpha$ factor from (0.2, 0.5, 0.8).  
  We report the tuned training size and the $\alpha$ factor that yields the highest F1 score in Table~\ref{table:hyperpara_ksize} and Table~\ref{table:hyperpara_alpha}.

  \begin{table*}[h]
    \small
    \centering
    
\begin{tabular}{ l rr c rrr } 
 
 \toprule
 
   & \multicolumn{2}{c}{Existing Benchmarks} & & \multicolumn{3}{c}{Our Benckmarks} \\ 

\cmidrule(lr){2-3} \cmidrule(lr){5-7}
   
   & Circa & SWDA\_IA & & Tennis & Movie & Air   \\ 
 
\midrule

\multicolumn{7}{l}{RoBERTa, training with} \\

\multicolumn{7}{l}{~~~in-domain instances and}\\

~~~~~~Circa         & n/a & n/a  & & 2k & 2k & 10k \\
~~~~~~SwDA-IA       & n/a & n/a  & & 2k & 2k & 5k \\
~~~~~~Circa+SwDA-IA & n/a & n/a  & & 2k & 2k & 10k \\

~~~all additional instances and & \\
~~~~~~Circa         & 5k  & 2k  & & 5k  & 5k & 2k  \\
~~~~~~SwDA-IA       & 2k  & 2k  & & 5k  & 2k & 10k \\
~~~~~~Circa+SwDA-IA & 10k & 10k & & 10k & 5k & 10k \\

\bottomrule

\end{tabular}
    \caption{Number of additional instances used in training. 
              We report the number (in thousands) that yields the highest F1 score. This table complements the third block of Table~\ref{t:interpretation_results}.}
    \label{table:hyperpara_ksize}
    
    \end{table*}

  \begin{table*}[h]
    \small
    \centering
    
\begin{tabular}{ l rr c rrr } 
 
    \toprule
    
      & \multicolumn{2}{c}{Existing Benchmarks} & & \multicolumn{3}{c}{Our Benckmarks} \\ 
   
   \cmidrule(lr){2-3} \cmidrule(lr){5-7}
      
      & Circa & SWDA\_IA & & Tennis & Movie & Air   \\ 
    
   \midrule

   RoBERTa, blended training with\\
   
   \multicolumn{7}{l}{~~~in-domain instances and}\\
   
   ~~~~~~Circa         & n/a & n/a & & 0.5 & 0.2 & 0.5 \\
   ~~~~~~SwDA-IA       & n/a & n/a & & 0.5 & 0.8 & 0.8 \\
   ~~~~~~Circa+SwDA-IA & n/a & n/a & & 0.2 & 0.2 & 0.5 \\
   
   ~~~all additional instances and&\\
   
   ~~~~~~Circa         & 0.5 & 0.5 & & 0.2 & 0.2 & 0.8 \\
   ~~~~~~SwDA-IA       & 0.2 & 0.2 & & 0.2 & 0.8 & 0.8 \\
   ~~~~~~Circa+SwDA-IA & 0.2 & 0.5 & & 0.5 & 0.8 & 0.8 \\

   \bottomrule
   \end{tabular}
    \caption{Tuned Blending factor ($\alpha$). 
    We report the $\alpha$ factor that yields the highest F1 score. This table complements the fourth block of Table~\ref{t:interpretation_results}.}
    \label{table:hyperpara_alpha}
    
    \end{table*}

\paragraph{Additional Results and Metrics with RoBERTa}

To better interpret our experimental results, we report results (F1 score) per label in Table~\ref{t:roberta_results_detailed}
and results with additional metrics (Precision, Recall, and F1 score) in Table~\ref{t:roberta_results_mutlimetric}.
These results complement Table~\ref{t:interpretation_results}.

\begin{table*}[t!]
  \small
  \centering
  
\begin{tabular}{ l rrrr c rrrrr  } 
 
 \toprule
 

   & \multicolumn{4}{c}{Circa} && \multicolumn{4}{c}{SWDA-IA}   \\ 
   
   \cmidrule(lr){2-5} \cmidrule(lr){7-10}

   & Yes & No & Mid & All && Yes & No & Mid & All  \\ 
\midrule

Majority Baseline & 0.74 & 0.00 & 0.00 & 0.43 && 0.00 & 0.00 & 0.66 &0.32  \\

\midrule

\multicolumn{7}{l}{RoBERTa, training with} \\

~~~~Circa         & 0.95 & 0.92 & 0.50 & 0.93 && 0.72 & 0.25 & 0.12 & 0.51  \\
~~~~SwDA-IA       & 0.75 & 0.64 & 0.15 & 0.69 && 0.77 & 0.52 & 0.29 & 0.63  \\
~~~~Circa+SwDA-IA & 0.95 & 0.93 & 0.38 & 0.93 && 0.79 & 0.57 & 0.44 & 0.68  \\
\midrule

\multicolumn{7}{l}{RoBERTa, training with} \\

\multicolumn{7}{l}{~~~in-domain instances and}\\

~~~~~~Circa         & \multicolumn{4}{c}{------------n/a------------} && \multicolumn{4}{c}{------------n/a------------}    \\
~~~~~~SwDA-IA       & \multicolumn{4}{c}{------------n/a------------} && \multicolumn{4}{c}{------------n/a------------}   \\
~~~~~~Circa+SwDA-IA & \multicolumn{4}{c}{------------n/a------------} && \multicolumn{4}{c}{------------n/a------------}    \\

~~~all additional instances and & \\
~~~~~~Circa         & 0.95 & 0.92 & 0.41 & 0.93  && 0.76 & 0.45 & 0.00 & 0.56   \\
~~~~~~SwDA-IA       & 0.82 & 0.68 & 0.13 & 0.74  && 0.80 & 0.56 & 0.32 & 0.67   \\
~~~~~~Circa+SwDA-IA & 0.95 & 0.93 & 0.43 & 0.93  && 0.79 & 0.61 & 0.47 & 0.70   \\
\midrule

RoBERTa, blended training with\\

\multicolumn{7}{l}{~~~in-domain instances and}\\

~~~~~~Circa         & \multicolumn{4}{c}{------------n/a------------} && \multicolumn{4}{c}{------------n/a------------} \\
~~~~~~SwDA-IA       & \multicolumn{4}{c}{------------n/a------------} && \multicolumn{4}{c}{------------n/a------------} \\
~~~~~~Circa+SwDA-IA & \multicolumn{4}{c}{------------n/a------------} && \multicolumn{4}{c}{------------n/a------------} \\

~~~all additional instances and&\\

~~~~~~Circa         & 0.95 & 0.93 & 0.51 & 0.93 && 0.74 & 0.37 & 0.05 & 0.54  \\
~~~~~~SwDA-IA       & 0.80 & 0.61 & 0.04 & 0.71 && 0.77 & 0.51 & 0.35 & 0.64  \\
~~~~~~Circa+SwDA-IA & 0.95 & 0.92 & 0.48 & 0.93 && 0.79 & 0.61 & 0.41 & 0.68  \\

\bottomrule
\end{tabular}


\noindent\begin{tabular}{ l rrrr@{\hspace{0.2cm}} rrrr@{\hspace{0.2cm}} rrrr } 

   \addlinespace
  
   \toprule
   

      & \multicolumn{4}{c}{Tennis} & \multicolumn{4}{c}{Movie} & \multicolumn{4}{c}{Air}   \\ 
     
     \cmidrule(lr){2-5} \cmidrule(lr){6-9} \cmidrule(lr){10-13}

     & Yes & No & Mid & All & Yes & No & Mid & All &  Yes & No & Mid & All \\ 
  \midrule
  
  Majority Baseline & 0.63 & 0.00 & 0.00 & 0.34     & 0.00 & 0.00 & 0.70 & 0.36     & 0.95 & 0.00 & 0.00 & 0.84 \\
  
  \midrule
  
  \multicolumn{7}{l}{RoBERTa, training with} \\
  
  ~~~~Circa         & 0.62 & 0.38 & 0.13 & 0.40     & 0.50 & 0.50 & 0.21 & 0.34     & 0.91 & 0.26 & 0.00 & 0.84 \\
  ~~~~SwDA-IA       & 0.61 & 0.33 & 0.23 & 0.42     & 0.49 & 0.29 & 0.34 & 0.37     & 0.70 & 0.07 & 0.00 & 0.63 \\
  ~~~~Circa+SwDA-IA & 0.72 & 0.55 & 0.25 & 0.52     & 0.53 & 0.43 & 0.27 & 0.37     & 0.93 & 0.17 & 0.00 & 0.84 \\
  \midrule
  
  \multicolumn{7}{l}{RoBERTa, training with} \\
  
  \multicolumn{7}{l}{~~~in-domain instances and}\\
  
  ~~~~~~Circa         & 0.70 & 0.42 & 0.07 & 0.43   & 0.55 & 0.46 & 0.02 & 0.24   & 0.92 & 0.30 & 0.00 & 0.85 \\
  ~~~~~~SwDA-IA       & 0.68 & 0.45 & 0.05 & 0.41   & 0.57 & 0.41 & 0.15 & 0.31   & 0.92 & 0.27 & 0.00 & 0.84 \\
  ~~~~~~Circa+SwDA-IA & 0.67 & 0.49 & 0.22 & 0.48   & 0.60 & 0.43 & 0.18 & 0.34   & 0.93 & 0.21 & 0.00 & 0.85 \\
  
  ~~~all additional instances and & \\
  ~~~~~~Circa         & 0.68 & 0.37 & 0.24 & 0.44   & 0.50 & 0.43 & 0.05 & 0.24   & 0.90 & 0.12 & 0.00 & 0.82  \\
  ~~~~~~SwDA-IA       & 0.70 & 0.46 & 0.27 & 0.51   & 0.59 & 0.42 & 0.07 & 0.28   & 0.90 & 0.17 & 0.00 & 0.82  \\
  ~~~~~~Circa+SwDA-IA & 0.67 & 0.50 & 0.36 & 0.53   & 0.57 & 0.47 & 0.33 & 0.42   & 0.92 & 0.22 & 0.00 & 0.84  \\
  \midrule
  
  RoBERTa, blended training with\\
  
  \multicolumn{7}{l}{~~~in-domain instances and}\\
  
  ~~~~~~Circa         & 0.65 & 0.38 & 0.00 & 0.37   & 0.59 & 0.42 & 0.13 & 0.31   & 0.90 & 0.24 & 0.00 & 0.83 \\
  ~~~~~~SwDA-IA       & 0.69 & 0.45 & 0.23 & 0.48   & 0.58 & 0.48 & 0.22 & 0.36   & 0.92 & 0.23 & 0.00 & 0.84  \\
  ~~~~~~Circa+SwDA-IA & 0.70 & 0.55 & 0.42 & 0.57   & 0.57 & 0.49 & 0.26 & 0.39   & 0.92 & 0.15 & 0.00 & 0.84  \\
  
  ~~~all additional instances and&\\
  
  ~~~~~~Circa         & 0.63 & 0.36 &  0.19 & 0.42   & 0.53 & 0.48 & 0.38 & 0.44   & 0.90 & 0.18 & 0.00 & 0.82 \\
  ~~~~~~SwDA-IA       & 0.69 & 0.47 &  0.27 & 0.50   & 0.62 & 0.41 & 0.29 & 0.40   & 0.93 & 0.30 & 0.00 & 0.85 \\
  ~~~~~~Circa+SwDA-IA & 0.69 & 0.55 &  0.49 & 0.59   & 0.59 & 0.42 & 0.48 & 0.49   & 0.94 & 0.24 & 0.00 & 0.86 \\

  \bottomrule
  \end{tabular}

  \caption{Detailed results (F1 score) obtained with RoBERTa per label. These results complement Table~\ref{t:interpretation_results}.}
  \label{t:roberta_results_detailed}
  \end{table*}

  \begin{table*}[t!]
    \small
    \centering
    
    \begin{tabular}{ l rrr r rrr  } 
 
        \toprule
        

          & \multicolumn{3}{c}{Circa} && \multicolumn{3}{c}{SWDA-IA}   \\ 
          
          \cmidrule(lr){2-4} \cmidrule(lr){6-8}

          & P & R & F1  && P & R & F1  \\ 
       \midrule
       
       Majority Baseline & 0.34 & 0.59 & 0.43 && 0.24 & 0.49 & 0.32   \\
       
       \midrule
       
       \multicolumn{7}{l}{RoBERTa, training with} \\
       
       ~~~~Circa         & 0.93 & 0.93 & 0.93 && 0.52 & 0.58 & 0.51    \\
       ~~~~SwDA-IA       & 0.69 & 0.70 & 0.69 && 0.65 & 0.65 & 0.63   \\
       ~~~~Circa+SwDA-IA & 0.93 & 0.93 & 0.93 && 0.68 & 0.69 & 0.68  \\
       \midrule
       
       \multicolumn{7}{l}{RoBERTa, training with} \\
       
       \multicolumn{7}{l}{~~~in-domain instances and}\\
       
       ~~~~~~Circa         & \multicolumn{3}{c}{----------n/a----------} && \multicolumn{3}{c}{-----------n/a-----------}    \\
       ~~~~~~SwDA-IA       & \multicolumn{3}{c}{----------n/a----------} && \multicolumn{3}{c}{-----------n/a-----------}   \\
       ~~~~~~Circa+SwDA-IA & \multicolumn{3}{c}{----------n/a----------} && \multicolumn{3}{c}{-----------n/a-----------}    \\
       
       ~~~all additional instances and & \\
       ~~~~~~Circa         & 0.93 & 0.93 & 0.93 && 0.67  & 0.63 & 0.56    \\
       ~~~~~~SwDA-IA       & 0.74 & 0.75 & 0.74 && 0.66  & 0.68 & 0.67    \\
       ~~~~~~Circa+SwDA-IA & 0.93 & 0.93 & 0.93 && 0.70  & 0.71 & 0.70    \\
       \midrule
       
       RoBERTa, blended training with\\
       
       \multicolumn{7}{l}{~~~in-domain instances and}\\
       
       ~~~~~~Circa         & \multicolumn{3}{c}{----------n/a----------} && \multicolumn{3}{c}{-----------n/a-----------} \\
       ~~~~~~SwDA-IA       & \multicolumn{3}{c}{----------n/a----------} && \multicolumn{3}{c}{-----------n/a-----------} \\
       ~~~~~~Circa+SwDA-IA & \multicolumn{3}{c}{----------n/a----------} && \multicolumn{3}{c}{-----------n/a-----------} \\
       
       ~~~all additional instances and\\
       
       ~~~~~~Circa         & 0.92 & 0.93 & 0.93 && 0.57 & 0.62 & 0.54   \\
       ~~~~~~SwDA-IA       & 0.72 & 0.72 & 0.71 && 0.65 & 0.66 & 0.64   \\
       ~~~~~~Circa+SwDA-IA & 0.93 & 0.93 & 0.93 && 0.69 & 0.71 & 0.68   \\

       \bottomrule
       \end{tabular}
       
       
       \noindent\begin{tabular}{ l rrr c rrr c rrr } 
       
          \addlinespace
         
          \toprule
          

             & \multicolumn{3}{c}{Tennis} && \multicolumn{3}{c}{Movie} && \multicolumn{3}{c}{Air}   \\ 
            
            \cmidrule(lr){2-4} \cmidrule(lr){6-8} \cmidrule(lr){10-12}

            & P & R & F1 && P & R & F1 &&  P & R & F1 \\ 
         \midrule
         
         Majority Baseline & 0.25 & 0.47 & 0.34      && 0.29 & 0.55 & 0.36      && 0.80 & 0.90 & 0.84 \\
         
         \midrule
         
         \multicolumn{7}{l}{RoBERTa, training with} \\
         
         ~~~~Circa         & 0.53 & 0.47 & 0.40     && 0.60 & 0.41 & 0.34       && 0.84 & 0.85 & 0.84 \\
         ~~~~SwDA-IA       & 0.61 & 0.33 & 0.42     && 0.53 & 0.40 & 0.37       && 0.83 & 0.53 & 0.63 \\
         ~~~~Circa+SwDA-IA & 0.60 & 0.58 & 0.52     && 0.57 & 0.40 & 0.37       && 0.83 & 0.86 & 0.84 \\
         \midrule
         
         \multicolumn{7}{l}{RoBERTa, training with} \\
         
         \multicolumn{7}{l}{~~~in-domain instances and}\\
         
         ~~~~~~Circa         & 0.69 & 0.51 & 0.43     && 0.72 & 0.34 & 0.24    && 0.84 & 0.86 & 0.85 \\
         ~~~~~~SwDA-IA       & 0.70 & 0.52 & 0.41     && 0.73 & 0.38 & 0.31    && 0.84 & 0.85 & 0.84 \\
         ~~~~~~Circa+SwDA-IA & 0.60 & 0.53 & 0.48     && 0.60 & 0.41 & 0.34    && 0.84 & 0.88 & 0.85 \\
         
         ~~~all additional instances and & \\
         ~~~~~~Circa         & 0.75 & 0.49 & 0.44     && 0.70 & 0.35 & 0.24    && 0.83 & 0.82 & 0.82  \\
         ~~~~~~SwDA-IA       & 0.65 & 0.57 & 0.51     && 0.58 & 0.36 & 0.28    && 0.84 & 0.81 & 0.82  \\
         ~~~~~~Circa+SwDA-IA & 0.67 & 0.57 & 0.53     && 0.59 & 0.46 & 0.42    && 0.22 & 0.00 & 0.84  \\
         \midrule
         
         RoBERTa, blended training with\\
         
         \multicolumn{7}{l}{~~~in-domain instances and}\\
         
         ~~~~~~Circa         & 0.30 & 0.48 & 0.37     && 0.59 & 0.39 & 0.31    && 0.83 & 0.82 & 0.83 \\
         ~~~~~~SwDA-IA       & 0.60 & 0.53 & 0.48     && 0.64 & 0.43 & 0.36    && 0.83 & 0.85 & 0.84  \\
         ~~~~~~Circa+SwDA-IA & 0.70 & 0.55 & 0.57     && 0.66 & 0.45 & 0.39    && 0.83 & 0.85 & 0.84  \\
         
         ~~~all additional instances and&\\
         
         ~~~~~~Circa         & 0.57 & 0.44 &  0.42    && 0.62 & 0.47 & 0.44    && 0.83 & 0.82 & 0.82 \\
         ~~~~~~SwDA-IA       & 0.66 & 0.54 &  0.50    && 0.68 & 0.45 & 0.40    && 0.84 & 0.87 & 0.85 \\
         ~~~~~~Circa+SwDA-IA & 0.60 & 0.60 &  0.59    && 0.62 & 0.50 & 0.49    && 0.84 & 0.88 & 0.86 \\

         \bottomrule
         \end{tabular}

    \caption{Results obtained with RoBERTa in Precision, Recall and F1 score. These results complement Table~\ref{t:interpretation_results}.}
    \label{t:roberta_results_mutlimetric}
    \end{table*}

\paragraph{Additional Results with BART}
    To minimize the variations by different models, we conduct experiments with BART-base using the same experimental setting as RoBERTa-base. 
    Table~\ref{table:bart_results} shows the results. Overall, BART underperforms RoBERTa on this task.

    \begin{table*}[t!]
      \small
      \centering
      
\begin{tabular}{ l rr c rrr } 
 
 \toprule
 
   & \multicolumn{2}{c}{Existing Benchmarks} & & \multicolumn{3}{c}{Our Benckmarks} \\ 

\cmidrule(lr){2-3} \cmidrule(lr){5-7}
   
   & Circa & SWDA-IA & & Tennis & Movie & Air   \\ 
 
\midrule

Majority Baseline & 0.43 & 0.32 & & 0.34 & 0.36 & 0.84 \\

\midrule

\multicolumn{7}{l}{BART, training with} \\

~~~~Circa         & 0.92 & 0.55 & & 0.41 & 0.27 & 0.84 \\
~~~~SwDA-IA       & 0.72 & 0.55 & & 0.43 & 0.35 & 0.83 \\
~~~~Circa+SwDA-IA & 0.92 & 0.66 & & 0.46 & 0.36 & 0.84 \\
\midrule

\multicolumn{7}{l}{BART, training with} \\

\multicolumn{7}{l}{~~~in-domain instances and}\\

~~~~~~Circa         & n/a & n/a  & & 0.45 & 0.24 & 0.78 \\
~~~~~~SwDA-IA       & n/a & n/a  & & 0.40 & 0.24 & 0.78 \\
~~~~~~Circa+SwDA-IA & n/a & n/a  & & 0.46 & 0.30 & 0.85 \\

~~~all additional instances and & \\
~~~~~~Circa         & 0.92  & 0.54  & & 0.35  & 0.32 & 0.83  \\
~~~~~~SwDA-IA       & 0.59  & 0.61  & & 0.43  & 0.39 & 0.86  \\
~~~~~~Circa+SwDA-IA & 0.92  & 0.63  & & 0.45  & 0.41 & 0.88  \\
\midrule

BART, blended training with\\

\multicolumn{7}{l}{~~~in-domain instances and}\\

~~~~~~Circa         & n/a & n/a & & 0.41 & 0.28 & 0.86 \\
~~~~~~SwDA-IA       & n/a & n/a & & 0.38 & 0.25 & 0.82 \\
~~~~~~Circa+SwDA-IA & n/a & n/a & & 0.49 & 0.37 & 0.86 \\

~~~all additional instances and&\\

~~~~~~Circa         & 0.92 & 0.52 & & 0.41 & 0.31 & 0.86 \\
~~~~~~SwDA-IA       & 0.58 & 0.65 & & 0.36 & 0.32 & 0.87 \\
~~~~~~Circa+SwDA-IA & 0.92 & 0.66 & & 0.52 & 0.42 & 0.88 \\

\bottomrule
\end{tabular}
      \caption{Results (F1 score) for interpreting indirect answers to yes-no questions with BART. 
      BART underperforms RoBERTa on this task.}
      \label{table:bart_results}
      \end{table*}

\paragraph{Experimental Details with LLMs}

Referring to Section~\ref{ssec:llm_results}, we test our benchmark with 
GPT-3.5 (\textit{gpt-35-turbo}), Alpaca (7B parameters), and Llama 2 (7B parameters). 
Figure~\ref{f:prompt} shows the prompts. 
For GPT-3.5, we call the API from Microsoft Azure. We set the \textit{temperature} to 0.1, \textit{top\_p} to 0.1, and \textit{max\_tokens} to 4 for optimal generation results.  
Both Alpaca and Llama are hosted locally using a single NVIDIA A100 (80GB) GPU.

\begin{figure}
  \begin{framed}
  \small

  \noindent\texttt{Below is an instruction and a yes-no question-answer pair input. Write a response that appropriately completes the request.}\\

  \noindent\texttt{\#\#\# Instruction: I need you to help me understand indirect answers to yes-no questions. 
  Indirect answers can be interpreted with three meanings: Yes, No, and Middle. 
  Simply reply Yes, No or Middle based on the question and answer.}\\
  
  \noindent\texttt{\#\#\# Input:}\\
  
  \noindent\texttt{Question: ``\textbf{<Question from benchmarks>}''}\\
  
  \noindent\texttt{Answer: ``\textbf{<Answer from benchmarks>}''}\\
  
  \noindent\texttt{Does the answer mean Yes, No or Middle?}\\
  
  \noindent\texttt{\#\#\# Response:}
  \end{framed}
  \caption{Prompts used with GPT, Alpaca, and Llama.}
  \label{f:prompt}
\end{figure}

\paragraph{Error Analysis on LLMs Results}

We conduct an error analysis for the results obtained with GPT-3.5 and the 4-shot prompt. Table~\ref{t:llm_error_analysis} lists the error distributions and errro examples.

\begin{table*}[t!]
  \small
  \centering
  

\begin{tabular}{ll rrr l }
    \toprule
    \multirow{3}{*}{Gold} & \multirow{3}{*}{Prediction} & \multicolumn{3}{c}{\%} & \multirow{3}{*}{Example} \\ 
    \cmidrule{3-5}
    &     &  Tennis & Movie & Air \\
    \midrule
    
    Y, N, M   & Fail to predict due to  &  1  & 25 & 3  & Q: Have you got to tell her your life story? \\
              & content filtering                    &&&& A: I'll say what I **** please.              \\ 
   \addlinespace
    Yes      &     No           &  21 & 2  & 3  & Q: Did you follow the Lance Armstrong stuff?   \\
                                            &&&&& A: A little bit.  \\  
    \addlinespace
    Yes      &     Middle       &  38 & 35 & 92 & Q: Are there any specifications?   \\
                                            &&&&& A: My departure time is evening.   \\ 
    \addlinespace
    No       &     Yes          &  0  & 1  & 0  & Q: Sure, do you prefer any class?   \\
                                            &&&&& A: I am ok with any class.  \\ 
    \addlinespace
    No       &     Middle       &  4  & 19 & 1  & Q: Are you working?   \\
                                            &&&&& A: Working? What do you mean, working? I'm walking.   \\
    \addlinespace 
    Middle   &     Yes          &  1  & 2  & 0  & Q: He went out and bought himself men's cologne\\   
                                            &&&&&  ~~~~~the other day. Did I tell you that?\\
                                            &&&&& A: Larry bought himself cologne?   \\ 
    \addlinespace
    Middle   &     No           &  35 & 16 & 1  & Q: Any reason why you felt you were down in the first \\
                                            &&&&& ~~~~~three sets in terms of quality?\\   
                                            &&&&& A: What's the question? \\

    \bottomrule
    
    \end{tabular}
  \caption{Error distributions with GPT-3.5 and the 4-shot prompt. The error percentages are categorized by Gold label and (wrong) Predictions.}
  \label{t:llm_error_analysis}
  \end{table*}

\end{document}